\documentclass[conference, 10pt]{IEEEtran}

\usepackage{amssymb, amsmath, epsfig}
\usepackage{graphicx}
\usepackage[ruled,vlined,linesnumbered]{algorithm2e}
\usepackage{bbm}
\usepackage{color}
\usepackage{caption}

\ifCLASSINFOpdf
\else
\fi

\newcommand{\bA}{\mathop{\boldsymbol A}}

\newcommand{\bC}{\mathop{\boldsymbol C}}

\newcommand{\bE}{\mathop{\boldsymbol E}}

\newcommand{\bI}{\mathop{\boldsymbol I}}

\newcommand{\bP}{\mathop{\boldsymbol P}}

\newcommand{\bX}{\mathop{\boldsymbol X}}
\newcommand{\bY}{\mathop{\boldsymbol Y}}
\newcommand{\bZ}{\mathop{\boldsymbol Z}}

\newcommand{\bc}{\mathop{\boldsymbol c}}

\newcommand{\be}{\mathop{\boldsymbol e}}

\newcommand{\bp}{\mathop{\boldsymbol p}}

\newcommand{\bs}{\mathop{\boldsymbol s}}

\newcommand{\bv}{\mathop{\boldsymbol v}}

\newcommand{\bx}{\mathop{\boldsymbol x}}
\newcommand{\by}{\mathop{\boldsymbol y}}
\newcommand{\bz}{\mathop{\boldsymbol z}}

\newcommand{\E}{\mathop{\mathbb E}}

\newcommand{\R}{\mathop{\mathbb R}}

\newtheorem{lemma}{Lemma}

\begin{document}
\title{Modeling Group Dynamics Using Probabilistic Tensor Decompositions}
\author{\IEEEauthorblockN{Lin Li, Ananthram Swami}
\IEEEauthorblockA{US Army Research Laboratory\\
Adephi, MD, USA}
\and
\IEEEauthorblockN{Anna Scaglione}
\IEEEauthorblockA{Electrical, Computer and Energy Engineering\\
Arizona State University\\
Tempe, AZ, USA}}

\maketitle
\begin{abstract}
We propose a probabilistic modeling framework for learning the dynamic patterns in the collective behaviors of social agents and developing profiles for different behavioral groups, using data collected from multiple information sources. The proposed model is based on a hierarchical Bayesian process, in which each observation is a finite mixture of an set of latent groups and the mixture proportions (i.e., group probabilities) are drawn randomly. Each group is associated with some distributions over a finite set of outcomes. Moreover, as time evolves, the structure of these groups also changes; we model the change in the group structure by a hidden Markov model (HMM) with a fixed transition probability. We present an efficient inference  method based on tensor decompositions and the expectation-maximization (EM) algorithm for parameter estimation. 
\end{abstract}
\IEEEpeerreviewmaketitle
\section{introduction}
In this paper, we consider the problem of modeling discrete social network data and learning the underlying group dynamics. The goal is to develop probabilistic profiles of large collections of data while preserving the essential temporal relationships that provide insights for various applications of interest. For example, in social network analysis, we want to analyze relationships between social agents and their behaviors over time and on various social media sites (i.e., Facebook, Twitter, Instagram, Google+, etc.).  In web advertising analysis, we want to analyze the relationships between customers and the types of products they buy from different shopping sites to capture customers\rq{} buying behaviors and learn the intrinsic factors that effect their buying decision process. In the study of scientific collaboration, using co-authorship networks from multiple journals on related subjects, one can analyze relationships between subjects and authors. This will, in turn, allow us to identify individual authors' area of expertise and potential research directions.

In the past few years, much effort has been devoted in the literature to develop models and algorithms that can efficiently extract the hidden groups/communities of a society. Generally speaking, there are two types of learning approaches. One approach starts by constructing networks that describe relationships or interactions between agents in a social setting; see~\cite{Fortunato201075}. The proposed algorithms focus on partitioning the graph into disjoint or overlapping subgraphs to reflect some of the shared features of the groups. 
The second approach focuses on extracting relevant information from data. For instance, the classical K-means clustering \cite{Lloyd1982,Banerjee2005,Chaudhuri_McGregor_2008} seeks to partition columns of a data matrix by minimizing the sum of squared distances of the data points to their respective cluster centroids. Co-clustering \cite{Banerjee2007} seeks to simultaneously partition rows and columns of a data matrix to form coherent groups, also known as co-clusters. Moreover, the study of multi-way clustering goes beyond pairwise relationship; it seeks to find group structures in a multi-dimensional array via tensor decompositions \cite{ICML2011Nickel_438}. Recently, tensor decompositions have also been applied to statistical relational learning as a new approach to infer the structure of a probabilistic graphical model \cite{Sutskever_modellingrelational, ICML2011Nickel_438, Song2013}. The latent Dirichlet allocation (LDA) model is another statistical approach for modeling collections of discrete data, such as document classification. It assumes that the words in a document are generated based on a mixture model. Specifically, it emphasizes that each documents contain multiple topics and the words in the document were drawn from these topics in proportion to the topic distribution of the document.

It is worth noting that some of the above-mentioned algorithms for extracting group information assume that  the data are from a single source and most of these algorithms do not deal with the problem when the group structure changes over time. Thus, the work that we discuss in this paper focuses on developing a modeling framework for {\it dynamically} fusing data from multiple related information sources, as well as learning the group dynamics. 

There are several reasons why one should be interested in extracting group information from multiple sources and characterizing patterns of temporal dynamics in different groups. First, in today\rq{}s highly computerized environment, large amounts of data are created every second from multiple sources that often share certain dependencies (e.g., mobile phones, SMS, GPS locator, emails, tweets, social networking sites, etc.). It makes sense to fuse data from all relevant information sources for joint analysis. Second, information collected from multiple sources can facilitate in accurately inferring the latent structure of the data.  Indeed, many machine learning tasks, such as classification, regression and clustering, can significantly improve their performance if information from multiple sources can be property integrated. Finally, understanding the dependency patterns over time and across multi-source data can be extremely beneficial in many science and engineering applications. In the context of social network analysis, each group can be identified by its membership and the specific actions that the members are likely to perform over time. Then the group probabilities provide an explicit representation of each social media site, which in turn, helps us determine the distances between these social media sites. In the study of scientific collaboration, one can analyze relationships between subjects and authors to identify groups with different research focus. Contrary to the graph-based model, 
each author may probabilistically belong to several communities (or subject group). Intuitively, an author's contribution to a community is proportional to the number of publications that he/she has published in these groups. However, estimating contribution by only publication count may be misleading. For example, it is quite possible that one has published less papers in one community turns out to have more contribution in that subject group because of the intrinsic characteristics of the group, such as the rate of publication, number of authors in the group, etc.

\subsection{Contributions}
While the main focus of most of the existing works is to extract group information from a single source in a static model, it is also important to consider fusing information from multiple information sources and over time. Our previous work \cite{Li7032228} provides a probabilistic modeling framework for extracting dynamic patterns from multi-source data. This paper is an extension of the work and it incorporates an efficient filter-based expectation-maximization (EM) learning method for performing parameter estimation on both synthetic data and real data. Specifically, we model the dynamics of a group by a hidden Markov model (HMM) and demonstrate how one can make use of a higher-order {\it tensor} \cite{Tucker1966, Bro1997, Kolda2009} (i.e., a multi-dimensional array) as an appropriate mathematical abstraction for the probabilistic graphical model \cite{Pearl1988, Ghosh1999, Jordan2004} that represents the conditional independence structure between latent variables. 

\subsection{Paper Organization}
The rest of the paper is organized as follows.
In Section \ref{sec:review}, we review the connection between graphical models and tensor decompositions, and explain how tensor decomposition can be used to transform a graphical model into a structurally simpler inference model. In Section \ref{sec:clustering}, we extend the model and formulate a dynamic tensor decomposition model for learning the low-dimensional structure from observed high-dimensional multi-source data. In Section \ref{sec:application}, we show the performance of the proposed framework for modeling group dynamics via simulations. 

\section{Mapping a Graphical Model into a Low-rank Tensor Decomposition: A Static Model}\label{sec:review}
In this section, groups and basic group features are defined. Their relationship with the LDA model is explained, followed by a description of how information from multiple information sources can be used to predict intrinsic group structure via tensor decomposition.

\subsection{Latent Groups}
Suppose there are $I$ information sources, i.e., $\Omega = \{a_1,\cdots,a_I\}$. Examples include various social network websites in social network analysis, different shopping sites in the analysis of web advertisement, different subjects/journals in the study of scientific collaboration, etc.. Given each source, we observe joint occurrences of two events $X$ and $Y$, with discrete outcomes in the finite sets $\mathcal{X} = \{x_1, \cdots, x_K\}$ and $\mathcal{Y} = \{y_1, \cdots, y_N\}$, respectively. One example is that of a person posting a picture. Here event $X$ is that a person is active and event $Y$ is that a picture is posted. Other examples include a customer buying a particular product, a researcher collaborating with another researcher, etc.. For simplicity, we only consider the case of observing joint occurrences of two events $X$ and $Y$. However, the approach can be generalized to a larger number of events with a higher dimensions. 

Let the tensor $\underline{\bP} \in {\R}^{K \times N \times I}$ be such that its $(k,n,i)^{\rm th}$ element denotes the joint probability that events $(x_k,y_n)$ are observed in the $i^{\rm th}$ information source:
\begin{align}\label{eqn:tensor}
P_{kni} = p(X = x_k, Y=y_n \mid a_i)\enspace.
\end{align}
Given $a_i$, if two events $X$ and $Y$ are independent, then
$
P_{kni} = p(X=x_k\mid a_i) p(Y=y_n \mid a_i)$. 

In practice, however, events $X$ and $Y$, conditioned on a particular information source $a_i$, are often dependent. For instance, in computing the probability of an agent posting a photo on Facebook, the agent and the particular activity (i.e., posting a photo) are two dependent variables. Similar arguments can be made for the shopping behavior of a customer and co-authoring a paper on a specific subject.

\begin{figure}[h!]
\centering
  \includegraphics[height = 17mm,width=27mm]{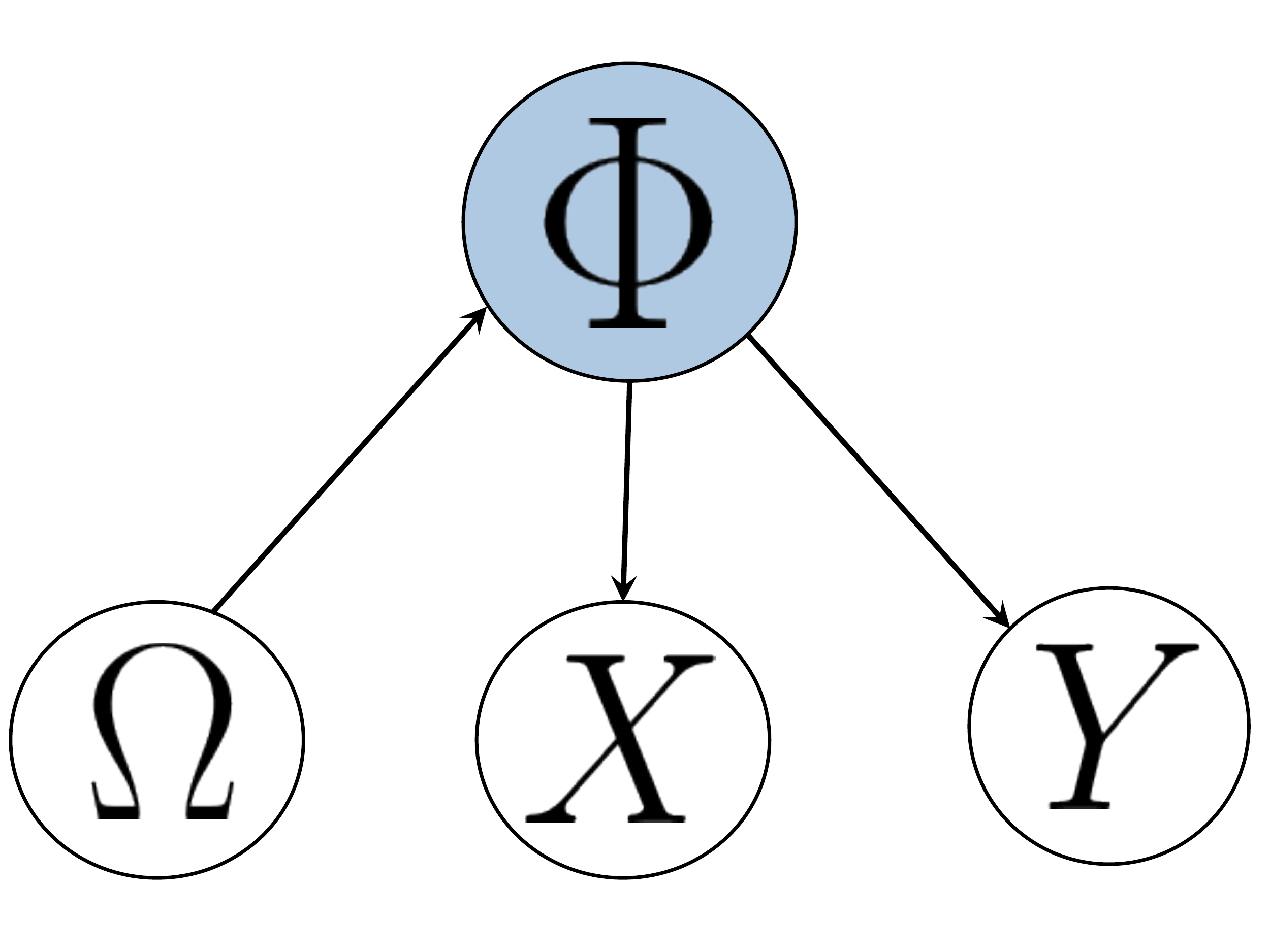}\vspace{-0.2cm}
\caption{Graphical representation of the tensor $\underline{\bP}$}\label{fig:StaticModel}
\end{figure}

Given that $X$ and $Y$ are dependent, let us assume that the information source itself is an event, determined by a latent variable $\Phi$, drawn from the groups  $\{\phi_1, \cdots, \phi_J\}$ such that conditioned on the groups, $X$ and $Y$ are independent. That is, probability mass functions (PMFs) of $X$ and $Y$ are conditionally independent given the group $\phi_i$:
$
p(x_k\mid a_i, \phi_j) = p(x_k\mid \phi_j)$ and 
$p(y_n \mid  a_i, \phi_j) = p(y_n \mid \phi_j)$.
Subsequently, Eqn. (\ref{eqn:tensor}) becomes
\begin{align}
P_{kni}&=\sum_{j=1}^Jp(\phi_j\mid a_i)p(x_k,y_n\mid a_i,\phi_j)\nonumber\\
&=\sum_{j=1}^Jp(\phi_j\mid a_i)p(x_k\mid \phi_j)p(y_n\mid \phi_j)\label{tensor_element}
\end{align}
where $\{p(\phi_j\mid a_i)\}_{j=1}^J$ represent group probabilities for source $a_i$ and they sum up to $1$. Here we assume that they are drawn randomly from a fixed distribution. Fig. \ref{fig:StaticModel} shows its corresponding graphical model.

Indeed, the above description is analogous to the LDA model, a generative model describing the generation of words in a collection of documents. That is, for a given information source $a_i$, a group distribution is drawn from a fixed distribution. (In the LDA model, this fixed distribution is modeled as a Dirichlet distribution). Let $n_i$ be the number of joint events to be sampled. For each of the $n_i$ samples, one of the $J$ groups is probabilistically drawn from the group probabilities $p(\phi_j\mid a_i)$. Given the group $\phi_j$, events $X$ and $Y$ are drawn independently from distributions $p(X|\phi_j)$ and $p(Y|\phi_j)$ respectively. 

Several simplifying assumptions are made in this model. First, the number of groups $J$ is assumed known and fixed. Second, the group features are represented by the probability vectors ${\bx}_j \in {\R}^K$ and ${\by}_j \in {\R}^N$, where ${\bx}_j(k) = p(x_k\mid \phi_j)$ and ${\by}_j(n) = p(y_n \mid \phi_j)$, respectively. For now, we treat them as fixed quantities for the given group.  Lastly, we treat $n_i$ as an ancillary variable and its value is generally known for a given dataset. Our goal is to estimate the latent groups from the $I$ information sources and the tensor decomposition offers a way to estimate these group features.

\subsection{Tensor Decompositions}
Given the group features ${\bx}_j \in {\R}^K$ and ${\by}_j \in {\R}^N$ for $j = 1, \cdots, J$, it follows from (\ref{tensor_element}) that the $i^{\rm th}$ {\it frontal} slice of the tensor $\underline{\bP}$ can be written as
\begin{align}\label{eqn:static}
{\bP}_{i} = \sum_{j=1}^J{c}_{ij}{\bx}_j{\by}_j^T\enspace,\enspace i = 1, \cdots ,I.
\end{align}
where ${c}_{ij} =  p(\phi_j| a_i)$ denotes the group probability. Furthermore, the above expression is equivalent to the following PARAFAC tensor decomposition: let ${\bc}_j \in {\R}^I$ be the $j^{\rm th}$ column of the matrix ${\bC} := [c_{ij}]$, then the given tensor $\underline{\bP}$ can be decomposed into
\begin{align}\label{tensor_decomposition}
\underline{\bP} = \sum_{j=1}^J{\bx}_j \circ {\by}_j \circ {\bc}_j
\end{align}
under the constraint that ${\boldsymbol 1}^T{\bx}_j=1$, ${\boldsymbol 1}^T{\by}_j=1$ and ${\bC}{\boldsymbol 1}={\boldsymbol 1}$,  where the symbol $\circ$ denotes the outer-product. 

\section{Clustering Behaviors in Dynamic Models}\label{sec:clustering}
The static model in (\ref{eqn:static}) and (\ref{tensor_decomposition}) assumes that for any given group $\phi_j$, the PMFs of $X$ and $Y$ are fixed quantities. Information from multiple sources can be fused together to extract group features using the tensor decomposition.  However, in many applications, group features often evolve over time; thus it is also important to understand how the group patterns change over time and across different information sources.  


\subsection{A Dynamic Model}
We model the dynamics of each information source $a_i$ by a Markov model and assume the following generative process for each information source $a_i$:
\begin{enumerate}
\item Choose group probabilities ${\boldsymbol c}_{i,:}=[c_{i1}, c_{i2}, \cdots, c_{iJ}]$ from some distribution
\item For each timestamp $t = 1, \cdots, T$:
\begin{enumerate}
\item Choose the number of joint events to be sampled: $n_i(t) \sim \operatorname{Poisson}(\omega)$
\item Choose a Markov state ${\bs}_{i,j}(t)$ from the distribution $p({\bs}_{i,j}(t)| {\bs}_{i,j}(t-1),{\bA}_{j})$ for all $j$
\item For each of the $n_i(t)$ sampled joint events $(X_\ell, Y_\ell)$ where $\ell = 1, \cdots, n_i(t)$:
\begin{enumerate}
\item Choose a group $\Phi_\ell \sim \operatorname{Multinomial}\left([c_{ij}]_{j=1}^J\right)$
\item Choose events $X_\ell$ and $Y_\ell$ independently from multinomial distributions $p(X_\ell \mid {\bX}_{\Phi_\ell}, {\bs}_{i,\Phi_\ell})$ and $p(Y_\ell \mid {\bX}_{\Phi_\ell}, {\bs}_{i,\Phi_\ell})$, respectively.
\end{enumerate}
\end{enumerate}
\end{enumerate} 

Several assumptions are made. First, the number of groups $J$ and the number of Markov states in each group are assumed known and fixed. Second, the group features evolve over time and we model it as a Markov process with a fixed group transition matrix ${\bA}_j$. Finally, the Poisson assumption for generating $n_i(t)$ is not critical to parameter estimation and its value is generally known given a dataset. 
 
\subsection{Hidden Markov Models}
Assume that the group probability $c_{ij} = p(\phi_j| a_i)$ is nonzero, at each time $t$, we associate $a_i$ to a state ${\bs}_{i,j}(t) \in \{{\be}_1, \cdots, {\be}_{Q_j}\}$ in a {\it hidden Markov model} (HMM), where ${\be}_q$ denotes the canonical base vector (with one in the $q^{\rm th}$ component and zero elsewhere) and $Q_j$ denotes the total number of states. Given the past states, the Markov property of a state transition implies
\begin{align*}
p\left({\bs}_{i,j}(t)|{\bs}_{i,j}(1), \cdots, {\bs}_{i,j}(t-1)\right) = p\left({\bs}_{i,j}(t)|{\bs}_{i,j}(t-1)\right)
\end{align*} 
i.e., the conditional distribution of the state at time $t$ depends only on the state at the previous time $t-1$ and not on the sequence of states that preceded it. For all $t \ge 0$ the transition probability from state ${\be}_q$ to state ${\be}_r$ is time-invariant. The matrix ${\bA}_j \in {\R}^{Q_j \times Q_j}$ with ${\bA}_j(q,r) = p\left({\bs}_{i,j}(t) = {\be}_r|{\bs}_{i,j}(t-1)={\be}_q\right)$ defines the transition matrix of the Markov chain and ${\bA}_j$ is row-stochastic. Moreover, given the state ${\bs}_{i,j}(t-1)$, the probability distribution of the state ${\bs}_{i,j}(t)$ can be written as
\begin{align}\label{eqn:Markov}
{\E}[{\bs}_{i,j}(t)\mid {\bs}_{i,j}(t-1)] = {\bA}_j^T{\bs}_{i,j}(t-1)\enspace.
\end{align}
Define ${\bv}_{i,j}(t):={\bs}_{i,j}(t) - {\bA}_j^T{\bs}_{i,j}(t-1)$. It follows from (\ref{eqn:Markov}) that the expectation of ${\bv}_{i,j}(t)$ conditioned on the past states equals zero:
${\E}[{\bv}_{i,j}(t)| {\bs}_{i,j}(1), \cdots, {\bs}_{i,j}(t-1)] = {\E}[{\bs}_{i,j}(t) - {\bA}_j^T{\bs}_{i,j}(t-1)| {\bs}_{i,j}(t-1)]=0$. This property will be used in a later section to derive a recursive state estimation filter.

The particular choice of using the coordinate vectors to represent
states in a HMM is attractive 
for several reason: 1) any function of the state can be
written as a linear function of coordinate vectors (and
hence it commutes with the expectation operator); 2) the
expected value of ${\bs}_{i,j}(t)$ equals the probability distribution
of the state, i.e., $\E\{\bs_{i,j}(t)\} = [p({\bs}_{i,j}(t) = {\be}_1), \cdots , p({\bs}_{i,j}(t) = {\be}_{Q_j} )]^T$.
This simplifies the state estimation of the Markov chain; see Section \ref{sec:learning} for a detailed discussion. 

\subsection{Tensor Representations}
\begin{figure}
\centering
  \includegraphics[height = 40mm,width=65mm]{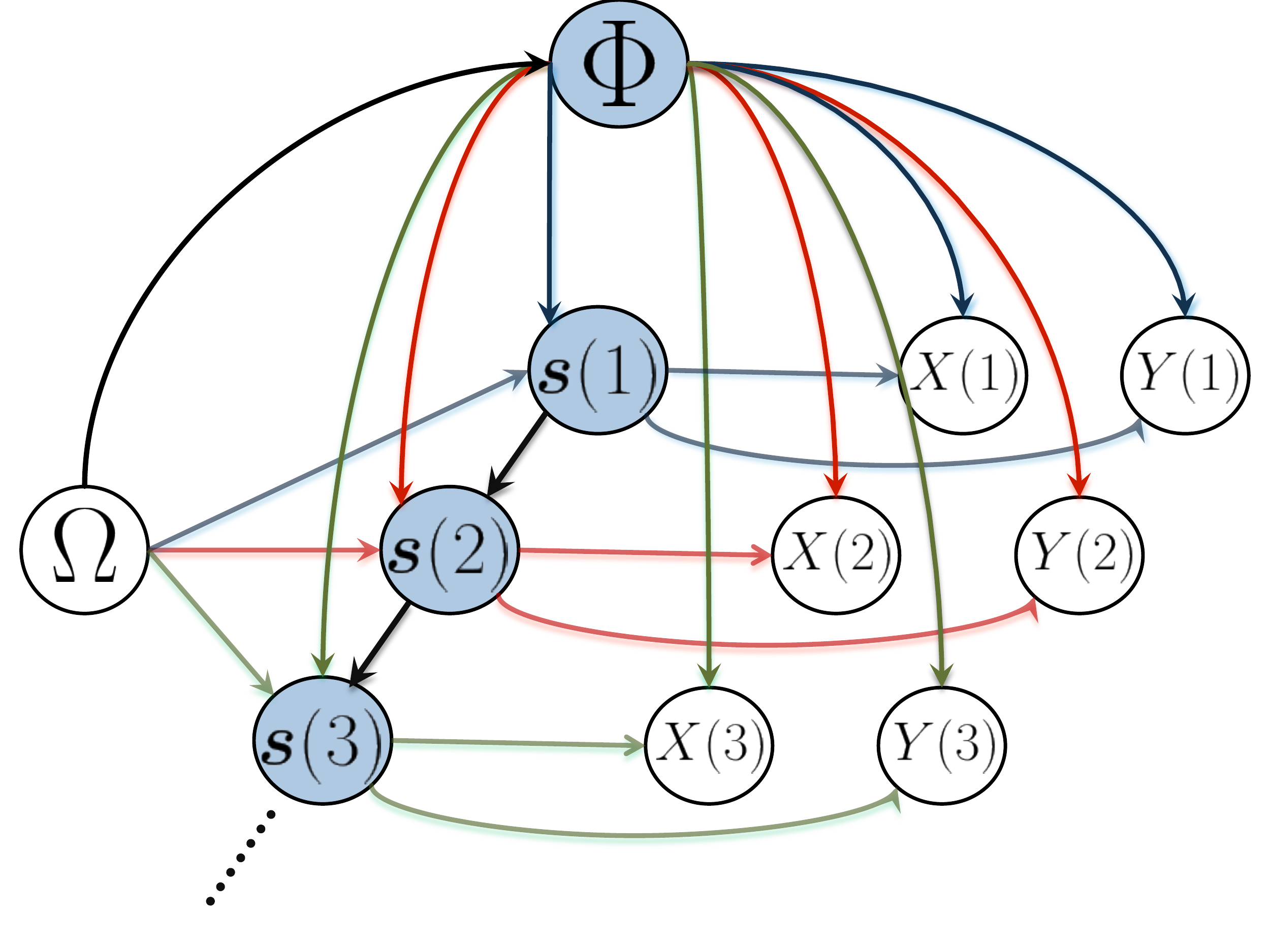}\vspace{-0.2cm}
\caption{Graphical representation of the dynamic tensor $\underline{\bP}(t$)}\label{fig:DynamicModel}
\end{figure}

Under the proposed dynamic model, given a group $\phi_j$, the transition probability ${\bA}_j$ is independent of $a_i$ and there exist two dictionaries of PMFs: ${\bX}_j = [{\bx}_{j,1}, \cdots, {\bx}_{j,Q_j}]$ and ${\bY}_j = [{\by}_{j,1}, \cdots, {\by}_{j,Q_j}]$ for the two random variables $X(t)$ and $Y(t)$, respectively; the $q^{\rm th}$ column of ${\bX}_j$ and ${\bY}_j$ correspond to the PMFs of $X(t)$ and $Y(t)$, respectively, when $a_i$ is in the state ${\bs}_{i,j}(t)={\be}_q$. Hence, the $i^{\rm th}$ frontal slice of the tensor $\underline{\bP}(t)$ obeys the following expression:
\begin{align} 
{\bP}_i(t) &= \sum_{j=1}^J{c}_{ij}{\bX}_j{\bs}_{i,j}(t){\bs}_{i,j}^T(t){\bY}_j^T\label{eqn:slice}
\end{align}
Fig. \ref{fig:DynamicModel} shows the graphical model corresponding to the dynamic tensor $\underline{\bP}(t)$.
Importantly, in contrast to the static model in (\ref{eqn:static}) where the group features are identified by the fixed PMFs ${\bx}_j$ and ${\by}_j$, the dynamic model assumes that given $\phi_j$, the group features  dynamically have the same set of PMFs for the variables $X$ and $Y$ with the same transition probability. 

Define $\bar{\bC}_j(t) \in {\R}^{I \times Q_j}$ to be the following matrix product:
\begin{align}\label{eqn:C}
\bar{\bC}_j(t) =\operatorname{diag}({\boldsymbol c}_j)[{\bs}_{1,j}(t), \cdots,{\bs}_{I,j}(t)]^T
\end{align}
 It follows from (\ref{eqn:slice}) that the decomposition of the time-varying tensor $\underline{\bP}(t)$ can be written as
\begin{align}
\underline{\bP}(t) &= \sum_{q = 1}^{Q_s}\bar{\bx}_{q} \circ \bar{\by}_{q} \circ \bar{\bc}_{q}(t)\enspace, \enspace Q_s=\sum_{j=1}^JQ_j\label{tensor}
\end{align}
where $\bar{\bx}_{q}$, $\bar{\by}_{q}$ and $\bar{\bc}_{q}(t)$ denote the $q^{\rm th}$ columns of matrices $\bar{\bX} = [{\bX}_1, \cdots, {\bX}_J]$,  $\bar{\bY} = [{\bY}_1, \cdots, {\bY}_J]$ and $\bar{\bC}(t) = [\bar{\bC}_1(t), \cdots, \bar{\bC}_J(t)]$, respectively. Note that the dynamics of the system is given by the $\bar{\bc}_q(t)$ vector, while the dictionaries of group properties are fixed. 

\subsection{Sampling Distribution: the Multinomial Case}
Recall from the generative process where at each time period, we sample $n_i(t)$ occurrences of the joint events. Let ${\boldsymbol \zeta}_i(t) \in {\R}^{K \times N}$ be a counting matrix whose $(k,n)^{\rm th}$ element represents the number of times the pair $(x_k, y_n)$ is observed from $a_i$ during the $t^{\rm th}$ time interval; $L_1$-norm of ${\boldsymbol \zeta}_i(t) $ equals $n_i(t)$. Moreover, the the counting matrix ${\boldsymbol \zeta}_i(t)$ follows a multinomial distribution with parameter ${\bP}_i$ and $n_i(t)$. 

Using the Central Limit Theorem \cite{Severini2012}, for $n_i(t)$  sufficiently large, one can approximate the multinomial distribution with the following multivariate normal distribution: 
$\frac{\operatorname{vec}({\boldsymbol \zeta}_{i}(t))}{n_i(t)} \sim \mathcal{N}\big({\bp}_i(t), \frac{1}{n_i(t)}\boldsymbol{\Sigma}_i(t)\big)$ where the mean is ${\bp}_i = \operatorname{vec}({\bP}_i)$ and the covariance is a function of the mean:
\begin{align}\label{Covariance}
\boldsymbol{\Sigma}_i(t)=\operatorname{diag}({\bp}_i(t))-{\bp}_i(t){\bp}_i^T(t).
\end{align}
Note that the covariance matrix $\boldsymbol{\Sigma}_i(t)$ is not full rank; the multivariate normal dsitribution is degerenate. We say that the normalized observation ${ \bz}_i(t):=\operatorname{vec}\left({\boldsymbol \zeta}_{i}(t)\right)/n_i(t)$ follows a {\it degenerate normal distribution}, whose density is given by
\begin{align}\label{Gaussian}
f\left({\bz}_i(t)\right) = \left|\frac{2 \pi \boldsymbol{\Sigma}_i(t)}{n_i(t)} \right|^{-\frac{1}{2}}_+e^{-\frac{n_i(t)}{2}\left({\bz}_i(t)-{\bp}_i(t)\right)^T\boldsymbol{\Sigma}^{+}_i(t)\left({\bz}_i(t)-{\bp}_i(t)\right)}
\end{align}
where $\boldsymbol{\Sigma}_i^+(t)$ is the generalized inverse of $\boldsymbol{\Sigma}_i(t)$ given in (\ref{Covariance}) and $|\cdot|_+$ denotes the pseudo-determinant. 

Importantly, computing the generalized inverse of the covariance matrxi $\boldsymbol{\Sigma}_i(t)$ can be computationally infeasible for large data sets. The following lemma gives explicit expressions for the generalized inverse and the pseudo-determinant. 

\begin{lemma}\label{lemma}
Given that $\bp$ is an $m$-dimensional probability vector (i.e., $\sum_{i=1}^mp_i = 1$ and $p_i \ge 0$), let  $\boldsymbol{\chi}_{>0}(\bp)$ be the indicator vector whose element equals $1$ if its corresponding element in $\bp$ is positive and $0$ otherwise. Let ${\bp}^+$ denote a $m$-dimentional vector such that its $i^{\rm th}$ element equals $1/p_i$ when $p_i > 0$ and 0 otherwise.
The pseudo-determinant and the generalized inverse of  $\boldsymbol{\Sigma} = \operatorname{diag}({\bp})-{\bp}{\bp}^T$ are given by 
\begin{align*}
|\boldsymbol{\Sigma}|_+ &=\|\boldsymbol{\chi}_{>0}(\bp)\|_1\prod_{i: {\bp}_i >0}{\bp}_i \\
\boldsymbol{\Sigma}^+ &= {\bC}\operatorname{diag}({\bp}^+) {\bC}^T
\end{align*}
where $\bC = \operatorname{diag}(\boldsymbol{\chi}_{>0}(\bp))\left({\bI}-\frac{\boldsymbol{1}\boldsymbol{1}^T}{\|\boldsymbol{\chi}_{>0}(\bp)\|_1}\right)$.
\end{lemma}

In summary, given the normalized observations ${\bZ}_i(t):={\boldsymbol \zeta}_i(t)/n_i(t)$, we have
\begin{align}
{\bZ}_i(t) &={\bP}_i(t)+{\bE}_i(t)\\
&= \sum_{j=1}^J{c}_{ij}{\bX}_j{\bs}_{i,j}(t){\bs}_{i,j}^T(t){\bY}_j^T+{\bE}_i(t)\nonumber
\end{align}
where the noise $\operatorname{vec}({\bE}_i(t)) \sim \mathcal{N}\big(0, \frac{1}{n_i(t)}\boldsymbol{\Sigma}_i(t)\big)$ is degenerate normal distributed; the matrix $\boldsymbol{\Sigma}_i(t)$ is a function of the mean vector ${\bp}_i(t)$.

 Note that $\underline{\bP}(t)$ is a function of the dictionaries ${\boldsymbol X}_j, {\boldsymbol Y}_j$, the group probabilties ${\bC}$, the transition matrices $\bA_j$ and the hidden states ${\bs}_{i,j}(t)$. They are the unknown parameters and can be estimated via the maximum likelihood techniques developed in \cite{Li_ITA, Li_PAMI}; see Section \ref{sec:MLE}. 

\section{Maximum Likelihood Inference}\label{sec:MLE}
Let $\boldsymbol{\xi}_{XYC}=\{{\boldsymbol X}_1, \cdots, {\boldsymbol X}_J, {\boldsymbol Y}_1, \cdots, {\boldsymbol Y}_J, {\bC}\}$ and $\boldsymbol{\xi}_A = \{{\boldsymbol A}_1, \cdots, {\boldsymbol A}_J\}$ be the sets of unknown parameters. Given the normalized observation tensors $\underline{\mathcal Z}_T = \{\underline{{\boldsymbol Z}}(1), \cdots, \underline{{\boldsymbol Z}}(T)\}$ and the count numbers $\{n_i(1), \cdots, n_i(T)\}$ for all $i$, the likelihood function $\mathcal{L}(\boldsymbol{\xi}_A, \boldsymbol{\xi}_{XYC})$ is given by
\begin{align}
\mathcal{L}&(\boldsymbol{\xi}_A, \boldsymbol{\xi}_{XYC}) = \sum_{{\bs}_{ij}(t) \forall i,j,t}
\prod_{t=1}^T\prod_{i=1}^Ip\big({\boldsymbol Z}_i(t)\mid {\bs}_{i,j}(t)~\forall j, \boldsymbol{\xi}_{XYC}\big)\nonumber \\ 
& \times\prod_{i=1}^I \prod_{j=1}^Jp\big({\boldsymbol s}_{i,j}(1)\big) \prod_{t=1}^{T-1}p\big({\boldsymbol s}_{i,j}(t+1)|{\boldsymbol s}_{i,j}(t),\boldsymbol{\xi}_A \big)
\end{align}
under the constraint that ${\bA}_j$, ${\bC}$ are non-negative row-stochastic and ${\bX}_j$, ${\bY}_j$ are non-negative column-stochastic.

For the ease of notation, let $\boldsymbol{\ell} =[\ell_1, \cdots, \ell_J]$ be an array of indices of the individual Markov states in their corresponding groups, i.e., ${\bs}_{i,j} = {\be}_{\ell_j}$ and $\ell_j \in \{1, \cdots, Q_j\}$. Let $({\boldsymbol \ell}_q)_{q=1}^Q$ with $Q = \prod_{j=1}^JQ_j$ be a {\it sequence} of the index array $\boldsymbol{\ell}$  corresponding to all the possible combinations of the Markov states; we denote $\{\boldsymbol{\ell}_q\}$ the set of all $\boldsymbol{\ell}_q$. We apply the {\it expectation-maximization} (EM) algorithm to find the maximum likelihood estimators by iteratively maximizing the following $\mathcal{Q}$-function: 
\begin{align*}
\mathcal{Q}(\boldsymbol{\xi}, \hat{\boldsymbol \xi}) =\sum_{i=1}^Ig_{i}(\boldsymbol{\xi}_A)+\sum_{i=1}^I\sum_{\boldsymbol{\ell}\in \{{\boldsymbol \ell}_q\}}h_i\big({\bp}_{i}^{\boldsymbol \ell}(\boldsymbol{\xi}_{XYC})\big)+r
\end{align*}
where $r$ is a constant and
\begin{align}
g_{i}(\boldsymbol{\xi}_A) =& \sum_{j=1}^J\sum_{\ell,k=1}^{Q_j}\log{\boldsymbol A}_j(\ell,k) \times \nonumber \\ &\mathbb{E}_{\hat{\boldsymbol \xi}}\bigg\{\sum_{t=1}^T\langle {\boldsymbol s}_{i,j}(t),{\boldsymbol e}_{\ell}\rangle \langle {\boldsymbol s}_{i,j}(t+1),{\boldsymbol e}_k \rangle \bigg| \underline{\mathcal Z}_T\bigg\}\\
h_i({\bp}_i^{\boldsymbol \ell}) =& \mathbb{E}_{\hat{\boldsymbol \xi}}\bigg\{\sum_{t=1}^T\prod_{j=1}^J\langle{\boldsymbol s}_{i,j}(t), {\boldsymbol e}_{\ell_j} \rangle  \log f\big(\boldsymbol{z}_i(t)\mid {\bp}_i^{\boldsymbol \ell}\big)\bigg|~\underline{\mathcal Z}_T \bigg\}
\end{align}
in which $\mathbb{E}_{\hat{\boldsymbol \xi}}\{\cdot\}$ denotes the expectation under the probability measure with model parameters equal to the estimated values (i.e., $\hat{\boldsymbol \xi}_A$ and $\hat{\boldsymbol \xi}_{XYC}$), $f\big(\boldsymbol{z}_i(t)\mid \bp_i^{\boldsymbol \ell}\big)$ is the probability density function given in (\ref{Gaussian}) and ${\bp}_i^{\ell}$ denotes the mean of the Gaussian observation ${\bz}_i$ from source $i$ given $\boldsymbol{\ell}$, that is,
\begin{align}
{\bp}_i^{\boldsymbol \ell} = \operatorname{vec}\bigg(\sum_{j=1}^J{c}_{ij}{\bX}_j{\be}_{\ell_j}{\be}_{\ell_j}^T{\bY}_j^T\bigg)\enspace.
\end{align}

\subsection{E-Step: Computing the $Q$-function}\label{sec:E_step}
It follows from the techniques in \cite{Li_ITA, Li_PAMI}, the following processes  are defined to compute the $Q$-function:
\begin{align}
\mathcal{J}_{i,j}^{k,m}(T) &= \sum_{t=1}^{T-1}\langle{\bs}_{i,j}(t),{\be}_k \rangle \langle{\bs}_{i,j}(t+1),{\be}_{m} \rangle\label{eqn:jump}\\
{\Gamma}_i^{\boldsymbol \ell}(T) &= \sum_{t=1}^T\prod_{j=1}^J\langle{\bs}_{i,j}(t),{\be}_{\ell_j} \rangle\label{eqn:combinedOccupation}\\
\mathcal{T}_i^{\boldsymbol \ell}(T, \boldsymbol{g}_i)&=\sum_{t=1}^T \prod_{j=1}^J\langle{\bs}_{i,j}(t),{\be}_{\ell_j}\rangle n_i(t) \boldsymbol{g}_i(t)\enspace.\label{eqn:mean}
\end{align}
In particular, the term $\mathcal{J}_{i,j}^{k,\ell}(T)$ defines the number of {\it jumps} from state $k$ to state $\ell$ for the source $i$ and the group $\phi_j$; $\Gamma_i^{\boldsymbol \ell}(T)$ defines the {\it combined occupation} time for source $i$ in the given Markov states associated with the index set ${\boldsymbol \ell}$. The term $\boldsymbol{g}_i(t)$ in Eqn. (\ref{eqn:mean}) takes one of the three forms:
\begin{align}
g_{i,1}(t) &= 1\\
\boldsymbol{g}_{i,2}(t) &= \boldsymbol{\chi}_{>0}({\bp}_{i}^{\boldsymbol \ell})\circ\boldsymbol{z}_i(t)\label{eqn:g_z}\\
\boldsymbol{g}_{i,3}(t) &= \boldsymbol{\beta}(t)\circ\boldsymbol{\beta}(t)
\end{align}
with $\boldsymbol{\beta}(t) =\operatorname{diag}(\boldsymbol{\chi}_{>0}({\bp}_{i}^{\boldsymbol \ell}))\left({\boldsymbol I}-\frac{\boldsymbol{1}{\boldsymbol 1}^T}{\|{\bp}_i^{\boldsymbol \ell}\|_0}\right)\boldsymbol{g}_{i,2}(t)$ and $\circ$ denotes the Hadamard product.

Given (\ref{eqn:jump}),  (\ref{eqn:combinedOccupation}) and (\ref{eqn:mean}), the E-step is reduced to the computation of the following conditional expectations: $\E_{\hat{\boldsymbol \xi}}\{\mathcal{J}_{i,j}^{k,m}(T)| \underline{\mathcal Z}_T\}$, $\E_{\hat{\boldsymbol \xi}}\{{\Gamma}_{i}^{\boldsymbol \ell}(T) | \underline{\mathcal Z}_T\}$ and $\E_{\hat{\boldsymbol \xi}}\{\mathcal{T}_{i}^{\boldsymbol \ell}(T,\boldsymbol{g}) | \underline{\mathcal Z}_T\}$.
See Section \ref{sec:learning} for the computation of the above expressions.

\subsection{M-step: Updating parameters in $\xi$}
The estimate of $\bA_j$ can be updated by maximizing $g_i({\boldsymbol \xi}_A)$, under the constraint that each ${\bA}_j$ is a row stochastic matrix. Using the method of Lagrange multipliers and equating the derivative to $0$ yields
\begin{align}\label{transition_matrix}
{\bA}_j(k,m) = \frac{\sum_{i=1}^I\E_{\hat{\boldsymbol \xi}}\{\mathcal{J}_{i,j}^{k,m}(T)\mid \underline{\mathcal Z}_T\}}{\sum_{i=1}^I\sum_{m}\E_{\hat{\boldsymbol \xi}}\{\mathcal{J}_{i,j}^{k,m}(T)\mid \underline{\mathcal Z}_T\}}\enspace.
\end{align}

 It follows from the $\mathcal{Q}$-function that the parameters in $\boldsymbol{\xi}_{\rm XYC}$ can be estimated by minimizing the following cost function:
\begin{align}
c(\boldsymbol{\xi}_{XYC}) = -\sum_{i=1}^I\sum_{\boldsymbol{\ell}\in \{{\boldsymbol \ell}_q\}}h_i\left({\bp}_i^{\boldsymbol \ell}({\boldsymbol \xi}_{XYC})\right). 
\end{align}
Gradient algorithm is used to update each of the parameters in the set $\boldsymbol{\xi}_{XYC}$.
 The gradient of the cost function with respective to any parameter $\xi \in \boldsymbol{\xi}_{XYC}$ can be written as
\begin{align}
\nabla c(\xi) = -\sum_{i=1}^I\sum_{{\boldsymbol \ell}\in \{{\boldsymbol \ell}_q\}}J({\bp}_i^{\boldsymbol \ell})(\xi)^T\nabla h_i({\bp}_i^{\boldsymbol \ell})
\end{align}
where $J({\bp}_i^{\boldsymbol \ell})(\xi)$ is the Jacobian matrix of ${\bp}_i^{\boldsymbol \ell}$ with respect to the parameter $\xi$ and 
\begin{align}\label{gradient}
\nabla h_i({\bp}_i^{\boldsymbol \ell}) &=  
\left(\frac{\boldsymbol{\bar b}_i^{\boldsymbol \ell}}{2}+\frac{\boldsymbol{\bar \zeta}_i^{\boldsymbol \ell}}{\|\boldsymbol{\chi}_{>0}({\bp}_i^{\boldsymbol \ell})\|_1}+{\kappa}_i^{\boldsymbol \ell}{\boldsymbol 1}\right)\circ({\bp}_i^{\boldsymbol \ell})^+\circ ({\bp}_i^{\boldsymbol \ell})^+\nonumber \\ &-\frac{\bar{a}_i^{\boldsymbol \ell}}{2}({\bp}_i^{\boldsymbol \ell})^+ -c_i^{\boldsymbol \ell}\boldsymbol{\chi}_{>0}({\bp}_{i}^{\boldsymbol \ell})
\end{align}
in which constants ${\kappa}_i^{\boldsymbol \ell}$ and $c_i^{\boldsymbol \ell}$ are given by
\begin{align}
{\kappa}_i^{\boldsymbol \ell} &=(\bar{n}_i^{\boldsymbol \ell}/2-\|\bar{\boldsymbol \zeta}_i^{\boldsymbol \ell}\|_1)/\|{\bp}_i^{\boldsymbol \ell}\|_0^2\\
c_i^{\boldsymbol \ell}&=(\bar{n}_i^{\boldsymbol \ell}-\|\bar{\boldsymbol \zeta}_i^{\boldsymbol \ell}\|_1)\frac{\|({\bp}_i^{\boldsymbol \ell})^+\|_1}{\|{\bp}_i^{\boldsymbol \ell}\|_0^2}+\frac{(\boldsymbol{\bar \zeta}_i^{\boldsymbol \ell})^T({\bp}_i^{\boldsymbol \ell})^+}{\|\boldsymbol{\chi}_{>0}({\bp}_i^{\boldsymbol \ell})\|_1}-\frac{\bar{n}_i^{\boldsymbol \ell}}{2}\enspace.
\end{align}
and $\bar{a}_i^{\boldsymbol \ell}$, $\bar{n}_i^{\boldsymbol \ell}$ $\bar{\boldsymbol \zeta}_i^{\boldsymbol \ell}$, $\bar{\boldsymbol b}_i^{\boldsymbol \ell}$ represent the following conditional expectations: 
\begin{align}
\bar{a}_i^{\boldsymbol \ell} &= {\E}_{\hat{\boldsymbol \xi}}\{{\Gamma}_i^{\boldsymbol \ell}(T)\mid \underline{\mathcal Z}_T\} \label{E_a}\\
\bar{n}_i^{\boldsymbol \ell} &= {\E}_{\hat{\boldsymbol \xi}}\{{\mathcal T}_i^{\boldsymbol \ell}(T, g_1)\mid \underline{\mathcal Z}_T\}\label{E_n}\\
\bar{\boldsymbol \zeta}_i^{\boldsymbol \ell} &={\E}_{\hat{\boldsymbol \xi}}\{{\mathcal T}_i^{\boldsymbol \ell}(T, \boldsymbol{g}_2(t))\mid \underline{\mathcal Z}_T\}\label{E_z}\\
\bar{\boldsymbol b}_i^{\boldsymbol \ell} &= {\E}_{\hat{\boldsymbol \xi}}\{{\mathcal T}_i^{\boldsymbol \ell}(T, \boldsymbol{g}_3(t))\mid \underline{\mathcal Z}_T\}\label{E_b}
\end{align}
The above conditional expectations can be computed in the $E$-step via filter-based learning, as discussed in the next section. 

\IncMargin{0.5em}
\begin{algorithm}[t!]
\SetAlgoNoLine
\SetKwData{Left}{left}\SetKwData{This}{this}\SetKwData{Up}{up}\SetKwFunction{Union}{union}\SetKwFunction{FindCompress}{FindCompress}
{\bf Input:} $\{\underline{\bZ}(1), \cdots, \underline{\bZ}(T)\}$ and $\{n_i(1), \cdots, n_i(T)\}$ for all $i = 1, \cdots, I$\;
{\bf Initialization:} $\{\hat{\bA}_j, \hat{\bX}_j, \hat{\bY}_j\}$ for all $j$ and $\hat{\bC}$\;  
\Indp
\Indm
\Repeat{some convergence criterion is met}{
{\it E-Step}: Compute $\E_{\hat{\boldsymbol \xi}}\{\mathcal{J}_{i,j}^{k,m}(T)\mid \underline{\mathcal Z}_T\}$ for all $(k,m)$\\
\Indp\Indp and $\bar{a}_i^{\boldsymbol \ell}$, $\bar{n}_i^{\boldsymbol \ell}$, $\bar{\boldsymbol \zeta}_i^{\boldsymbol \ell}$, $\bar{\boldsymbol b}_i^{\boldsymbol \ell}$ in (\ref{E_a}) - (\ref{E_b})\;
\Indm\Indm{\it M-Step}: Update parameters\\
\Indp\Indp $\hat{\bA}_j = \arg \max_{{\bA}_j}\sum_{i=1}^Ig_i({\bA}_j)$ using (\ref{transition_matrix})\;
$\hat{\bX}_j\leftarrow \mathcal{P}\left(({\boldsymbol 1}-\alpha_{X_j}\nabla c({\bX}_j))\circ\hat{\bX}_j\right)$\;
$\hat{\bY}_j\leftarrow \mathcal{P}\left(({\boldsymbol 1}-\alpha_{Y_j}\nabla c({\bY}_j))\circ\hat{\bY}_j\right)$\;
$\hat{\bC}\leftarrow \mathcal{P}\left(({\boldsymbol 1}-\alpha_C\nabla c({\bC}))^T\circ\hat{\bC}^T\right)^T$\;
}
\Indp\Indm\Return{}
\caption{Maximum Likelihood Estimation}\label{alg.MLE}
\end{algorithm}\DecMargin{0.5em}

Algorithm \ref{alg.MLE} summarizes the main steps for parameter estimation. The algorithm starts with initial estimates of the parameters $\{\hat{\bA}_j,\hat{\bX}_j, \hat{\bY}_j, \hat{\bC}\}$. The specific choice of the initial estimates and model order selection rule will be discussed in Section \ref{sec:application} where numerical results are reported. The (generalized) EM algorithm alternates between the E-step, which computes the set of conditional expectations in the $\mathcal{Q}$-function, and a M-step, which updates the parameters by maximizing the $\mathcal{Q}$-function. In particular, the transition matrices ${\bA}_j$ for all $j$ can be updated using (\ref{transition_matrix}). Since there does not exist a closed-form solution for parameters $\{{\bX}_j, {\bY}_j, {\bC}\}$, an iterative projected gradient algorithm is used. Specifically, $\mathcal{P}(\cdot)$ denotes the projection function that projects columns of the input matrix onto their corresponding canonical simplexes. The expression inside the projection function is a regular gradient descent step in which the stepsize is set to be sufficiently small and proportional to the parameters to be updated.

\section{Filter-Based Learning}\label{sec:learning}
In this section, we introduce a filter-based algorithm for learning the conditional expectation that are required in the evaluation of the $\mathcal{Q}$-function in the E-step. Contrary to the traditional forward-backward algorithm, the filter-based algorithm does not required extra storage for the intermediate quantities. The filtering processes for different terms in the E-step is independent, thus making parallel processing possible if needed. 

\subsection{Filter-Based State Estimation}
This section investigate the estimation of the state distribution ${\E}_{\hat{\boldsymbol \xi}}\{{\bs}_{i,j}(t)\mid \underline{\mathcal Z}_T\}$ for all $i=1, \cdots, I$ and $j = 1, \cdots, J$. The key is to introduce a measure transformation, which is defined by the likelihood ratio 
\begin{align}
\lambda_i^{\boldsymbol \ell}(t) = \frac{f({\bz}_i(t)\mid {\bp}_i^{\boldsymbol \ell})}{f_Q({\bz}_i(t))}
\end{align}
 where the numerator $f_{\boldsymbol \ell}({\bz}_i(t) | {\bp}_i^{\boldsymbol \ell})$ represents the probability distribution of the Gaussian observation ${\bz}_i(t) \sim \mathcal{N}\left({\bp}_i^{\boldsymbol \ell},{\boldsymbol \Sigma}^{\boldsymbol \ell}_i \right)$ given in (\ref{Gaussian}) and $f_Q({\bz}_i(t))$ denotes the probability distribution of the observation for ${\bz}_i(t) \sim \mathcal{N}\left(0,\big({\boldsymbol \Sigma}^{\boldsymbol \ell}_i\big)^+{\boldsymbol \Sigma}^{\boldsymbol \ell}_i \right)$. Thus, given ${\bp}_i^{\boldsymbol \ell}$ and the function $\boldsymbol{g}_{i,2}(t)$ in (\ref{eqn:g_z}), it follows from Lemma \ref{lemma} that the computation of the likelihood ratio is reduced to the following expression: 
\begin{align}\label{likelihoodRatio}
\lambda_i^{\boldsymbol \ell}(t) &= n_i(t)^{\frac{\|\boldsymbol{\chi}_{>0}({\bp}_i^{\boldsymbol \ell})\|_1-1}{2}}|\boldsymbol{\Sigma}_i^{\boldsymbol \ell}|_+^{-\frac{1}{2}}\exp\bigg(\frac{\|\boldsymbol{g}_{i,2}(t)\|_2}{2}-\nonumber\\
&\frac{\|\boldsymbol{g}_{i,2}(t)\|_1^2}{2\|\boldsymbol{\chi}_{>0}({\bp}_i^{\boldsymbol \ell})\|_1}
-\frac{n_i(t)}{2}\big[{\bz}_i(t)^T\big(({\bp}_i^{\boldsymbol \ell})^+\circ{\bz}_i(t)\big)-1-\nonumber\\
&2\eta_i^{\boldsymbol \ell}(t){\bz}_i(t)^T({\bp}_i^{\boldsymbol \ell})^++\left(\eta_i^{\boldsymbol \ell}(t)\right)^2\|({\bp}_i^{\boldsymbol \ell})^+\|_1\big]\bigg)
\end{align}
where $\eta_i^{\boldsymbol \ell}(t) = \frac{1-\|\boldsymbol{g}_{i,2}(t)\|_1}{\|\boldsymbol{\chi}_{>0}({\bp}_i^{\boldsymbol \ell})\|_1}$. 
Let
\begin{align}
\Lambda_{i}(t) = \prod_{\tau = 1}^t \lambda_i(\tau)
\end{align}
be the product of the transformation up to time $t$.  The conditional Bayes' Theorem yields
\begin{align}\label{bayes}
{\E}_{\hat{\boldsymbol \xi}}\{{\bs}_{i,j}(t)\mid \underline{\mathcal Z}_t\} &= \frac{\E_Q\{\Lambda_{i}(t){\bs}_{i,j}(t)\mid \underline{\mathcal Z}_t\}}{\E_{Q}\{\Lambda_{i}(t)\mid \underline{\mathcal Z}_t\}}\nonumber\\
&= \frac{\E_Q\{\Lambda_{i}(t){\bs}_{i,j}(t)\mid \underline{\mathcal Z}_t\}}{\big\langle {\E}_Q\{\Lambda_{i}(t){\bs}_{i,j}(t)\mid \underline{\mathcal Z}_t\},{\boldsymbol 1}\big\rangle}
\end{align}
Hence, given the current estimates $\hat{\boldsymbol \xi}$, the state distribution for source $i$ can be described by normalizing the vector $\E_Q\{\Lambda_{i}(t){\bs}_{i,j}(t)| \underline{\mathcal Z}_t\}$.
 Our goal  is to derived a recursive expression for computing this un-normalized state distribution.

Define $\mathcal{F}({\bs}_{i,j}(t)) = \E_Q\{\Lambda_{i}(t){\bs}_{i,j}(t)| \underline{\mathcal Z}_t\}$. Let $\mathcal{A}_{j}(m) \subset \{\boldsymbol{\ell}_q\}$ such that $\boldsymbol{\ell} \in \mathcal{A}_j(m)$ if and only if its $j^{\rm th}$ element of $\boldsymbol{\ell}$ equals $m$. Then one can derive the following recursive relation: (see Appendix) 
\begin{align}\label{state_filter}
\mathcal{F}({\bs}_{i,j}(t+1)) &= \operatorname{diag}\left(\boldsymbol{\omega}_{i,j}(t+1)\right){\bA}_j^T\mathcal{F}({\bs}_{i,j}(t)) \\
\text{where}\enspace\enspace \boldsymbol{\omega}_{i,j}(t) &= \left[{\sum}_{\boldsymbol{\ell}\in \mathcal{A}_j(1)}\lambda_i^{\boldsymbol{\ell}}(t), \dots, {\sum}_{\boldsymbol{\ell}\in \mathcal{A}_j(Q_j)}\lambda_i^{\boldsymbol{\ell}}(t)\right].\label{omega}
\end{align}
 Then it follows from (\ref{bayes}) that the probability distribution of the Markov state is given by
\begin{align}\label{state_estimation}
{\E}_{\hat{\boldsymbol \xi}}\{{\bs}_{i,j}(t)\mid \underline{\mathcal{Z}}_t\} = \frac{\mathcal{F}\left({\bs}_{i,j}(t)\right)}{\langle \mathcal{F}({\bs}_{i,j}(t)),{\boldsymbol 1}\rangle}
\end{align}
and  the estimated state of source $i$ in group $j$ at time $t$ is simply the one that corresponds to the highest likelihood.

Similarly, one can also compute the probability distribution of the set of Markov states. Given the sequence $(\boldsymbol{\ell}_q)_{q=1}^Q$ of index arrays and the transition matrices ${\bA}_j$ for all $j$, one can construct another hidden Markov model in which $\boldsymbol{\pi}_i(t) \in \{{\be}_1, \cdots, {\be}_Q\}$ denotes the state at time $t$ and $\Phi \in \R^{Q \times Q}$ denotes the tranition matrix.  Each state is associated to its corresponding element in the sequence $(\boldsymbol{\ell}_q)_{q=1}^Q$ and the $(k, m)^{\rm th}$ element of the transition matrix $\boldsymbol{\Phi}$ represents the probability of transitioning from state ${\boldsymbol \pi}(t) = {\be}_{k}$ to state ${\boldsymbol \pi}(t+1) = {\be}_{m}$. Let $\mathcal{F}({\boldsymbol \pi}_i(t)) = {\E}_Q\{\Lambda_i(t)\boldsymbol{\pi}_i(t)|\underline{\mathcal Z}_T\}$. Following a derivation similar to that in (\ref{state_filter}), it can be shown that
\begin{align}
{\E}_{\hat{\boldsymbol \xi}}\{\boldsymbol{\pi}_i(t)|\underline{\mathcal Z}_T\} = \frac{\mathcal{F}(\boldsymbol{\pi}_i(t))}{\langle \mathcal{F}(\boldsymbol{\pi}_i(t)), {\boldsymbol 1}\rangle}
\end{align} 
where $\mathcal{F}(\boldsymbol{\pi}_i(t))$ is given by the following recursive expression:
\begin{align}\label{recursive_pi}
\mathcal{F}({\boldsymbol \pi}_{i}(t+1))&=\operatorname{diag}\left(\boldsymbol{\lambda}_i(t+1)\right)\boldsymbol{\Phi}^T\mathcal{F}({\boldsymbol \pi}_{i}(t))\\
\text{where}\enspace\enspace
\boldsymbol{\lambda}_i(t) &= [\lambda_i^{\boldsymbol{\ell}_1}(t), \cdots, \lambda_i^{\boldsymbol{\ell}_Q}(t)].\label{lambda}
\end{align}
We are now ready to computer the conditional expectations in the $\mathcal{Q}$-function. 

\subsection{Filter-Based Computation of the $\mathcal{Q}$-Function}
Recall from Section \ref{sec:E_step} that the computation of the $\mathcal{Q}$-function is reduced to computing the following conditional expectations: $\E_{\hat{\boldsymbol \xi}}\{\mathcal{J}_{i,j}^{k,m}(T)| \underline{\mathcal Z}_T\}$, $\E_{\hat{\boldsymbol \xi}}\{{\Gamma}_{i}^{\boldsymbol \ell}(T) | \underline{\mathcal Z}_T\}$ and $\E_{\hat{\boldsymbol \xi}}\{\mathcal{T}_{i}^{\boldsymbol \ell}(T,\boldsymbol{g}) | \underline{\mathcal Z}_T\}$. The first term is used to estimate the transition matrix ${\bA}_j$ for all $j$ as given in~(\ref{transition_matrix}). The second  and the third terms the gradient of the cost function $C(\boldsymbol{\xi}_{XYC})$ with respect to the parameters in the set $\boldsymbol{\xi}_{XYC}$; see Eqn.~(\ref{gradient}) and (\ref{E_a}) - (\ref{E_b}).  In this section, we will discuss how to recursively compute these terms using the technique introduced in the previous section.

Define $\mathcal{F}(\mathcal{J}_{i,j}^{k,m}(t){\bs}_{i,j}(t)) = {\E}_Q\{\Lambda_i(t)\mathcal{J}_{i,j}^{k,m}(t){\bs}_{i,j}(t)|\underline{\mathcal Z}_t\}$. Then at time $T$, we have
\begin{align}
{\E}_{\hat{\boldsymbol \xi}}\{\mathcal{J}_{i,j}^{k,m}(T)|\underline{\mathcal Z}_t\} &= \langle {\E}_{\hat{\boldsymbol \xi}}\{\mathcal{J}_{i,j}^{k,m}(T){\bs}_{i,j}(T)|\underline{\mathcal Z}_T\}, {\boldsymbol 1}\rangle\nonumber\\
&= \frac{\langle \mathcal{F}(\mathcal{J}_{i,j}^{k,m}(T){\bs}_{i,j}(T)) , {\boldsymbol 1}\rangle}{\langle \mathcal{F}({\bs}_{i,j}(T)), {\boldsymbol 1}\rangle}
\end{align}
Note that the term $\mathcal{F}({\bs}_{i,j}(T))$  in the denominator can be computed recursively using (\ref{state_filter}). Following a similar procedure as in~(\ref{state_filter}), one can derive the following recursive expression:
\begin{align}\label{filter_J}
&\mathcal{F}(\mathcal{J}_{i,j}^{k,m}(t+1){\bs}_{i,j}(t+1))\\
=&\operatorname{diag}\left(\boldsymbol{\omega}_{i,j}(t+1)\right){\bA}_j^T\mathcal{F}(\mathcal{J}_{i,j}^{k,m}(t){\bs}_{i,j}(t))\nonumber\\
&+\boldsymbol{\omega}_{i,j}(t+1,m){\bA}_j(k,m)\langle \mathcal{F}({\bs}_{i,j}(t)),{\be}_k\rangle{\be}_m\enspace.\nonumber
\end{align}

Similarly, let $\mathcal{F}(\Gamma_i^{\boldsymbol \ell}(t){\boldsymbol \pi}_i(t))~=~{\E}_Q\{\Lambda_i(t)\Gamma_i^{\boldsymbol \ell}(t){\boldsymbol \pi}_i(t)|\underline{\mathcal Z}_t\}$ and $\mathcal{F}(\mathcal{T}_i^{\boldsymbol \ell}(t,g_r){\boldsymbol \pi}_i(t))~=~{\E}_Q\{\Lambda_i(t)\mathcal{T}_i^{\boldsymbol \ell}(t,g_r){\boldsymbol \pi}_i(t)|\underline{\mathcal Z}_t\}$, respectively, where $g_r$ denotes the $r^{\rm th}$ element of $\boldsymbol{g}$. Then for $\boldsymbol{\ell} = \boldsymbol{\ell}_q$, we have $\prod_{j=1}^J\langle{\bs}_{i,j}(t),{\be}_{\ell_j}\rangle = \langle{\boldsymbol \pi}_i(t),{\be}_q \rangle$. Thus it can be shown that
\begin{align}
{\E}_{\hat{\boldsymbol \xi}}\{\Gamma_{i}^{\boldsymbol{\ell}_q}(T)|\underline{\mathcal Z}_t\}
&= \frac{\langle \mathcal{F}(\Gamma_{i}^{\boldsymbol{\ell}_q}(T){\boldsymbol \pi}_{i}(T)) , {\boldsymbol 1}\rangle}{\langle \mathcal{F}({\boldsymbol \pi}_{i}(T)), {\boldsymbol 1}\rangle}\\
{\E}_{\hat{\boldsymbol \xi}}\{\mathcal{T}_{i}^{\boldsymbol{\ell}_q}(T,g^r)|\underline{\mathcal Z}_t\}
&= \frac{\langle \mathcal{F}(\mathcal{T}_{i}^{\boldsymbol{\ell}_q}(T,g^r){\boldsymbol \pi}_{i}(T)) , {\boldsymbol 1}\rangle}{\langle \mathcal{F}({\boldsymbol \pi}_{i}(T)), {\boldsymbol 1}\rangle}
\end{align}
where the term $\mathcal{F}({\boldsymbol \pi}_{i}(T))$ in the denominator is given by the recursive expression in~(\ref{recursive_pi}) and 
\begin{align}\label{filter_Gamma}
&\mathcal{F}(\Gamma_i^{\boldsymbol{\ell}_q}(t+1){\boldsymbol \pi}_i(t+1))\\
=&\operatorname{diag}\left(\boldsymbol{\lambda}_i(t+1)\right)\boldsymbol{\Phi}^T\mathcal{F}(\Gamma_i^{\boldsymbol{\ell}_q}(t){\boldsymbol \pi}_i(t))\nonumber\\
&+\boldsymbol{\lambda}_i(t+1,q)\boldsymbol{\Phi}(:,q)^T\mathcal{F}({\boldsymbol \pi}_i(t)){\be}_q\nonumber
\end{align}
\begin{align}\label{filter_T}
&\mathcal{F}(\mathcal{T}_i^{\boldsymbol{\ell}_q}(t+1,g^r){\boldsymbol \pi}_i(t+1))\\
=&\operatorname{diag}\left(\boldsymbol{\lambda}_i(t+1)\right)\boldsymbol{\Phi}^T\mathcal{F}(\mathcal{T}_i^{\boldsymbol{\ell}_q}(t,g^r){\boldsymbol \pi}_i(t))\nonumber\\
&+n_i(t+1)g^r(t+1)\boldsymbol{\lambda}_i(t+1,q){\boldsymbol \Phi}(:,q)^T\mathcal{F}({\boldsymbol \pi}_i(t)) {\be}_q\nonumber
\end{align}
Using the filters derived in (\ref{filter_J}), (\ref{filter_Gamma}), (\ref{filter_T}), one can then evaluate the $\mathcal{Q}$-function and perform parameter update, as shown in Algorithm \ref{alg.MLE}.

\section{Numerical Result}\label{sec:application}
In this section, we present numerical results using both synthetic data and real data. 
We use K-means clustering to initilaize the parameters. It is an extremely fast algorithm and has no hyper-parameter to tune beyond setting the model orders. 

\subsection{Synthetic Datasets}
Two groups $\phi_1$ and $\phi_2$ are generated, thus $J = 2$. The first group $\phi_1$ has $Q_1 = 2$ hidden Markov states with a fixed transition matrix ${\bA}_1 \in \R^{2 \times 2}$ while the second group has $Q_2 = 3$ hidden Markov states with a fixed transition matrix ${\bA}_2 \in \R^{3 \times 3}$. $I$ information sources are generated; each is associated with some group distribution $[c_{i1}, \cdots, c_{iJ}]$, uniformly sampled over a $(J-1)$-dimensional simplex. Moreover, for each latent group $\phi_j$, a pair of dictionaries ${\bX}_j^{6 \times Q_j}$ and ${\bY}_j^{8 \times Q_j}$ are generated; each column in ${\bX}_j$ and ${\bY}_j$ represents a discrete probability vector, such that the resulting tensor $\underline{\bP}(t)$ in (\ref{tensor}) at each time $t$ satisfies the sufficient condition for a unique decomposition  \cite{Kruskal1977,Stegeman2007540,Berge2002}.
Observations $\underline{\boldsymbol Z}(t)$ for $t = 1,\dots, T$ are generated according to the model described in Section \ref{sec:clustering}. In the simulation, we set the number of sampled events $n_i(t)$ to be a constant for all $t$ and all $i$, here we simply denote it as $n$. Recall that larger value of $n$ implies that observations are less noisy.

The performance of the learning algorithm is measured by the mean-squared-error (MSE) between the recovered (or estimated) tensors $\hat{\underline{\boldsymbol Z}}(t)$ and the observations $\underline{\boldsymbol Z}(t)$:
\begin{align*}
\text{MSE} = \frac{1}{IT}\sum_{t = 1}^T\sum_{i=1}^I\|\hat{{\boldsymbol Z}}_i(t) -\underline{\boldsymbol Z}_i(t)\|_2^2
\end{align*}

\begin{figure}[h!]
\centering
  \includegraphics[height = 30mm]{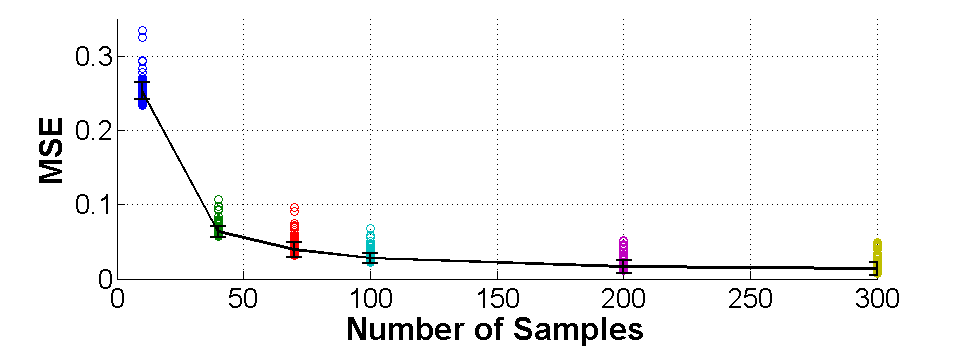}
\caption{Averaged MSE as the sample size increases for $n=5$ and $T=100$}\label{fig:scatter}
\end{figure}

\begin{figure}[h!]
\centering
  \includegraphics[height = 31mm]{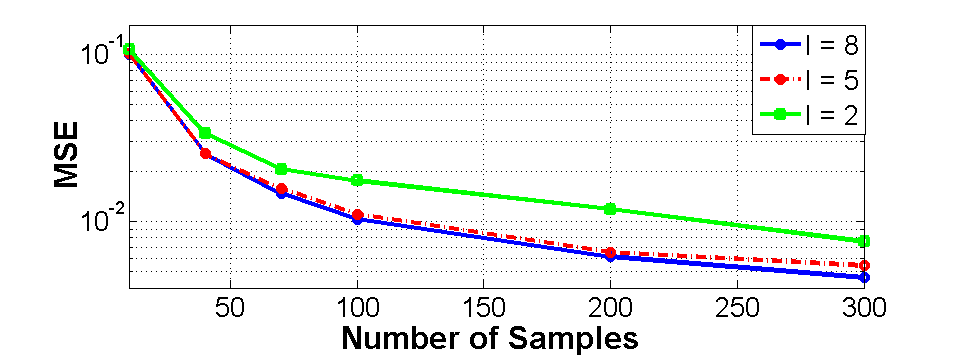}
\caption{Averaged MSE for $I = 2,5,8$ information sources}\label{fig:sources}
\end{figure}

\begin{figure}[h!]
\centering
  \includegraphics[height = 34mm]{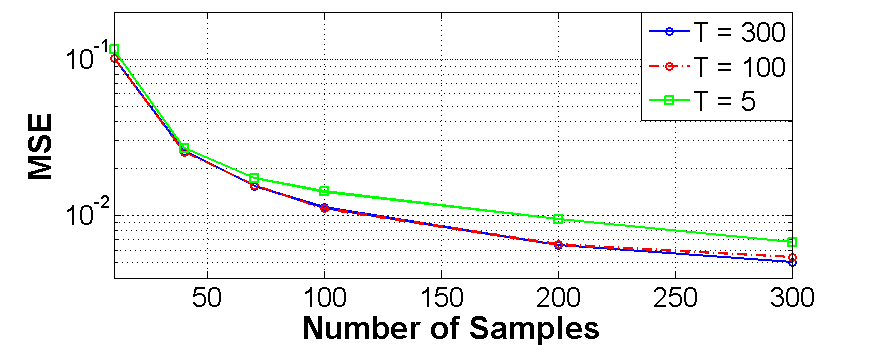}
\caption{Averaged MSE for $T = 5,100,300$ observations}\label{fig:times}
\end{figure}

Fig. \ref{fig:scatter} shows the performance of the learning algorithm for $I=5$ sources and $T = 100$ observation time as the sample size $n$ increase. For each value of $n$, the algorithm is run $200$ times; each with a different set of group probabilities while the group features $\bX$ and $\bY$ are fixed. Observe that the averaged MSE over $200$ runs decreases as the sample size increases. 
Fig. \ref{fig:sources} shows the performance of the learning algorithm for different number of sources, i.e., $I = 2,5,8$. Observe that the averaged MSE (over $200$ runs) decreases as the number of information sources increases.
 Fig. \ref{fig:times} show the performance of the learning algorithm for $T = 5,100,300$ observations. Observation that averaged MSE (over $200$ runs) decreases significantly from $T=5$ to $T = 100$, but there is hardly any improvement from $T=100$ to $T=300$. 

\subsection{Synthetic Datasets with Missing Values}
\begin{figure}[h!]
\centering
  \includegraphics[height = 35mm]{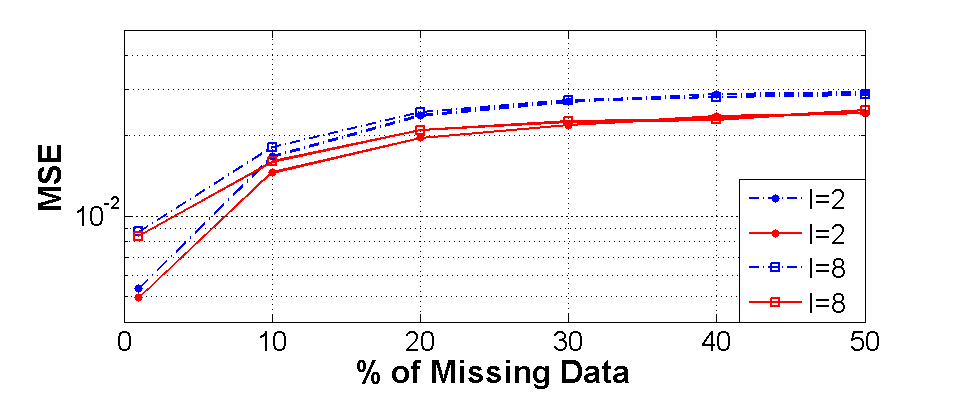}
\caption{Averaged MSE for $T=100$ and $n = 300$}\label{fig:missingdata}
\end{figure}

This section investigate the robustness of the proposed learning algorithm when there are missing data. Here we assume that the loss is uniformly random over time. The algorithm treats the missing data as zero. Fig. \ref{fig:missingdata} shows the averaged MSE (over $200$ runs) for $I =2, 8$ information sources as the percentage of missing data increases. The dotted lines represent the MSE between the recovered tensors and the observed tensors with missing values, while the solid lines represent the MSE between recovered tensors and the original tensors with no missing value. Observed that the recovered tensors are closer to the original tenors than the input tensors with missing values. Moreover, when the percentage of missing data is small, $I=8$ information sources performs better. As the percentage of missing data increases, the advantage of having more sources diminishes.

\begin{figure}[h!]
\centering
  \includegraphics[height = 35mm]{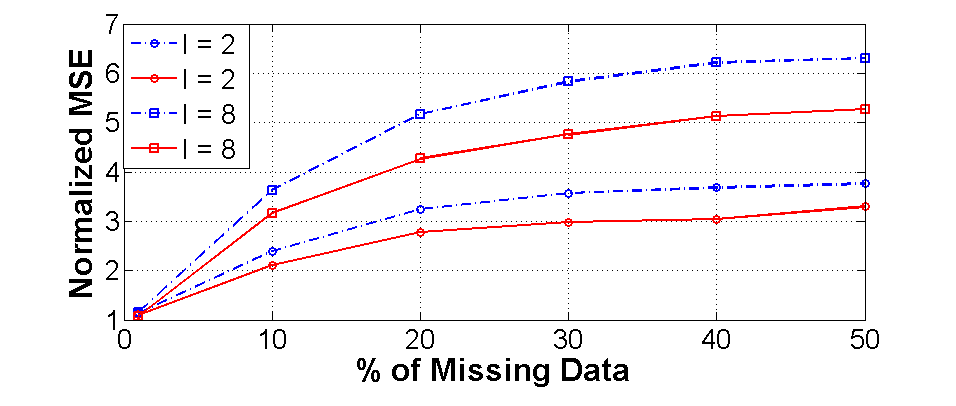}
\caption{Normalized MSE $T=100$ and $n = 300$}\label{fig:Ratio_missingdata}
\end{figure}

To get a clearer idea of the effects of missing data, Fig. \ref{fig:Ratio_missingdata} shows the normalized MSE as the percentage of missing data increases. Here we normalized the averaged MSE with missing data (Fig. \ref{fig:missingdata}) by the averaged MSE when there is no missing data. Again the dotted line represent the normalized MSE between the recovered tensors and the observed tensors with missing values, and the solid lines represent the normalized MSE between the recovered tensors and the original tensors with no missing value. Observe that the loss in performance is more significant for larger values of $I$.


\subsection{APS dataset: Atomic, Molecular and Condense Matter Physics}
In this section, we validate the proposed model using the bibliographic database from American Physical Society (APS). Five journals are considered, i.e., PRA, PRB, PRC, PRD and PRE, from the year $1986$ to the year $2009$. We selected three research areas for investigation: (1) Atomic and Molecular Physics; (2) Condensed Matter: Structural, Mechanical and Thermal Properties; (3) Condensed Matter: Electronic Structure, Electrical, Magnetic, and Optical Properties. There are a total of $23$ subjects under these research areas. We excluded the subjects that have less than $100$ publications for any given sample period because they are likely to increase the variability of the data and decrease the performance of the estimation. Table \ref{table:subjects} lists the resulting $14$ subjects and their corresponding Physics and Astronomy Classification Scheme (PACS). Furthermore, we excluded authors who have published less than $30$ papers over the years of interest, because authors who has published small number of papers do not provide us with sufficient samples of their co-authors over the years. 

\begin{center}
    \begin{tabular}{ | l | l | p{6cm} |}
    \hline
    ~ & PACS &  Subjects \\ \hline
    1 & 31 &  Electronic structure of atoms and molecules: theory \\ \hline
    2 & 32 & Atomic properties and interactions with photons. \\ \hline
    3 & 34 &  Atomic and molecular collision processes and interactions \\ \hline
4 & 61 &  Structure of solids and liquids; crystallography \\ \hline
    5 & 63 & Lattice dynamics \\ \hline
    6 & 64 &  Equations of state, phase equilibria, and phase transitions \\ \hline
7 & 68 &  Surfaces and interfaces; thin films and nanosystems (structure and nonelectronic properties) \\ \hline
    8 & 71 & Electronic structure of bulk materials \\ \hline
    9 & 72 & Electronic transport in condensed matter \\ \hline
10 & 73&  Electronic structure and electrical properties of surfaces, interfaces, thin films, and low-dimensional structures \\ \hline
    11 & 74 & Superconductivity\\ \hline
    12 & 75 &   Magnetic properties and materials \\ \hline
12 & 78 &  Optical properties, condensed-matter spectroscopy and other interactions of radiation and particles with condensed matter \\ \hline
    14 & 79 &Electron and ion emission by liquids and solids; impact phenomena \\ \hline
    \end{tabular}\label{table:subjects}
\captionof{table}{PACS and subjects}
\end{center}

The temporal tensor data are constructed as follows. Each subject is considered as one information source. For each subject and a given time block, we construct a co-authorship network, in which the nodes represent authors and two authors are connected if they have co-authored at least one paper. The strength of their connection is given by the number of papers they've collaborated on. Hence, the $i^{\rm th}$ slice of resulting tensor $\underline{\bZ}(t)$ at time $t$ is given by the weighted adjacency matrix of the corresponding co-authorship network, normalized by the sum of the weights of its edges, i.e., the total number of paper published in the respective subject. Each time $t$ represents a $3$-year period; thus there are a total of $T = 8$ time periods from year $1986$ to the year $2009$.

\subsubsection{Model Order Selection and Parameter Initialization}
To select an optimal number groups and the number of the Markov states within each group, we use the method proposed by Pham, et.al. in \cite{Pham2004}, which builds on the $k$-means algorithm. Specifically, we first compute the degree centralities of all the weighted graphes and use them as inputs to the method proposed in \cite{Pham2004} to select an optimal number of groups. The $k$-means algorithm is performed to give group labels to the input data points. For the data points of a given group label, we run the method in \cite{Pham2004} again to select an optimal number of Markov states for the given group and use $k$-means algorithm to initialize the group dictionary ${\bX}_j$ for all the groups $j = 1, \cdots, J$. Since the co-authorship network is un-directed, we have ${\bY}_j = {\bX}_j$. To initialize the group probability matrix $\bC$,  for each subject, we count the group labels of the degree centrality vectors for each group, and assign a higher probability to the group with a higher count number. To avoid getting stuck in a local minimum, several different probability matrices $\bC$ are tested initially and we pick the one that gives the smallest MSE.

\subsubsection{Numerical Results}
\begin{figure}[h!]
\centering
  \includegraphics[height = 35mm]{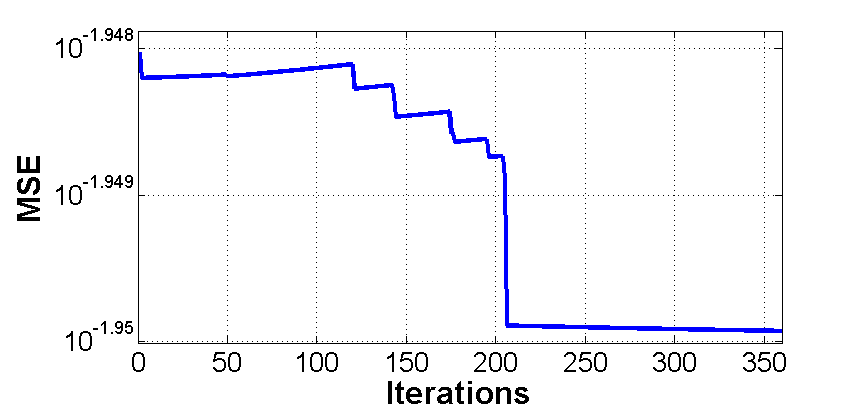}
\caption{Evolution of the MSE as the number of iteration increases}\label{fig:MSE}
\end{figure}

\begin{figure}[h!]
\centering
  \includegraphics[height = 55mm]{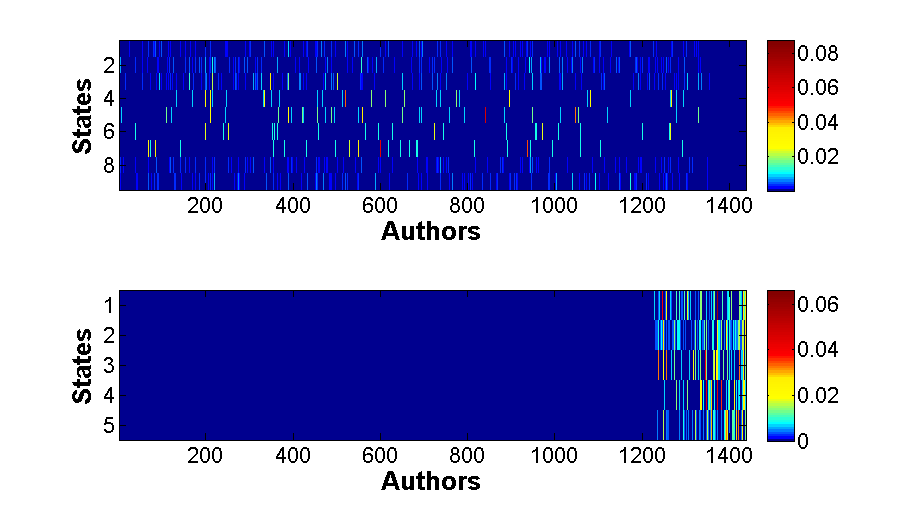}
\caption{Estimated group dictionary for the first group (top plot) and for the second group (bottom plot)}\label{fig:groups}
\end{figure}

\begin{figure}[h!]
\centering
  \includegraphics[height = 40mm]{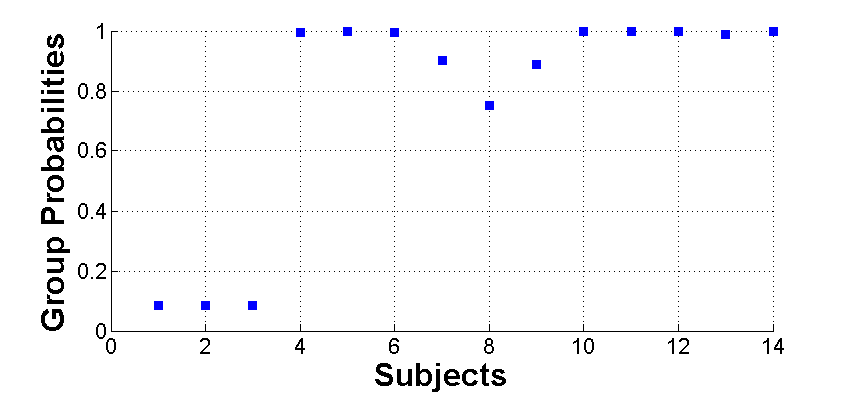}
\caption{Group probabilities}\label{fig:probabilities}
\end{figure}

Given the dataset as described earlier, the model selection procedure indicates that there are $2$ groups and the numbers of Markov states in the respective groups are $9$ and $5$. Fig. \ref{fig:MSE} shows the evolution of the MSE. Observe that the MSE, which measures the goodness of fit, decreases as the number of maximum likelihood iteration increases. 
Fig. \ref{fig:groups} shows the estimates the group dictionaries. The x-axis represents the authors, the y-axis represents discrete Markov states, and the color represents the participation of the authors given the state in a group. Observe that each group has a set of active authors. There are $128$ authors who belong to both groups, $1227$ authors who only publish in group $1$ and $83$ authors who only publish in group $2$. Fig. \ref{fig:probabilities} shows the group probability of each of the subjects belonging to group $1$. Observe that the first $3$ subjects are similar to each other and subjects $4-14$ are similar to each other. Indeed, it follows from Table \ref{table:subjects} that the first $3$ subjects are in the area of atomic and molecular physics while the rest of the subjects are in the area of condense matter. In addition, from the group probabilities shown in Fig. \ref{fig:probabilities}, it can be  seen that the two groups are very loosely related.  

\subsection{APS dataset: Elementary Particles and Nuclear Physics}
In this section, we investigate the fields of elementary particles and nuclear physics in the APS dataset. It is to be noted that the model order selection protocol returns $1$ group on all the subjects unders these two areas. That is, if we consider all the authors who have published more than $30$ papers under these two areas, we can not tell these subjects apart; they are closely related. 

To obtain a more insightful result, we now consider only the ``experts'' in the fields, that is, authors who have published more than $70$ papers. Then we eliminate the subjects that don't have sufficient number of publications in any given time period. The resulting subjects of interest are listed in Table \ref{table:subjects_particles}. PACS $11-14$ belong to the area of Elementary Particles and Fields, while PACS $21,23,27$ belong to the area of Nuclear Physics. The model order selection protocol suggests that there are two groups; the first group has $5$ Markov states and the second group has $8$ Markov states.
\begin{center}
    \begin{tabular}{ | l | l | p{6cm} |}
    \hline
    ~ & PACS &  Subjects \\ \hline
    1 & 11 &  Electronic structure of atoms and molecules: theory \\ \hline
    2 & 12 & Atomic properties and interactions with photons. \\ \hline
    3 & 13 &  Atomic and molecular collision processes and interactions \\ \hline
4 & 14 &  Structure of solids and liquids; crystallography \\ \hline
    5 & 21 & Lattice dynamics \\ \hline
    6 & 23 &  Equations of state, phase equilibria, and phase transitions \\ \hline
7 & 27 &  Surfaces and interfaces; thin films and nanosystems (structure and nonelectronic properties) \\ \hline
    \end{tabular}\label{table:subjects_particles}
\captionof{table}{PACS and subjects}
\end{center}

\begin{figure}[h!]
\centering
  \includegraphics[height = 38mm]{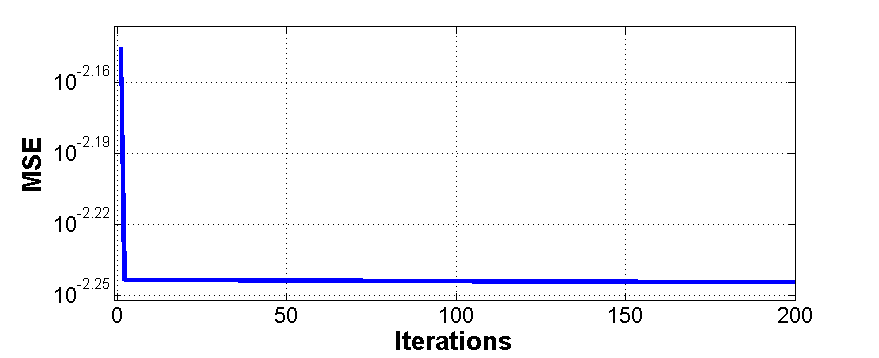}
\caption{Evolution of the MSE as the number of iteration increases}\label{fig:particles_MSE}
\end{figure}

\begin{figure}[h!]
\centering
  \includegraphics[height = 40mm]{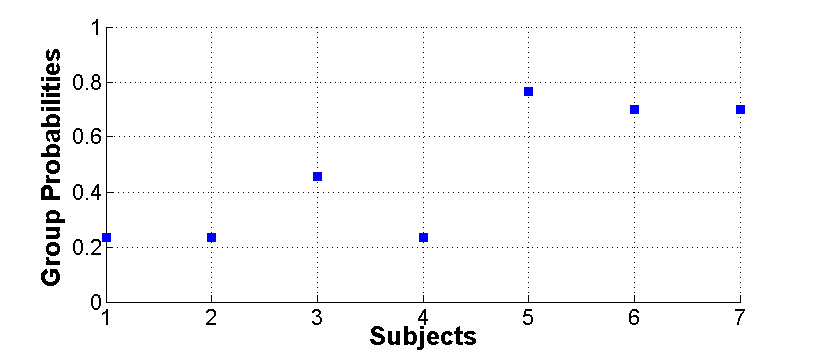}
\caption{Group probabilities}\label{fig:particles_probabilities}
\end{figure}

Fig. \ref{fig:particles_MSE} shows the evolution of the MSE as the number of iterations increases. Observe that the MSE decreases monotonically. Fig. \ref{fig:particles_probabilities} shows the estimated group dictionaries. The color represents individual authors participations' in their respective groups and for a given state. Observe that each group has a set of active authors. There are $52$ authors who only publish in the first group, $1335$ authors who only publish in the second group, and $24$ authors who publish in both groups. Fig. \ref{fig:particles_probabilities} shows the group probabilities of each subject belonging to group $1$. If the threshold for classification is set to be $0.5$, then it can be seen that the first four subjects form one cluster and they represents subjects in the area of Elementary Particles and Fields, while the last three subjects form the second cluster and they represent Nuclear Physics. In Contrast to Fig. \ref{fig:probabilities} in the previous section, the distances between the group probabilities of different subjects are closer. It implies that these subjects are very much related because authors who publish in the subjects in the area of Elementary Particles and Fields are also likely to publish in the subjects in the area of Nuclear physics. 

\begin{figure}[h!]
\centering
  \includegraphics[height = 70mm]{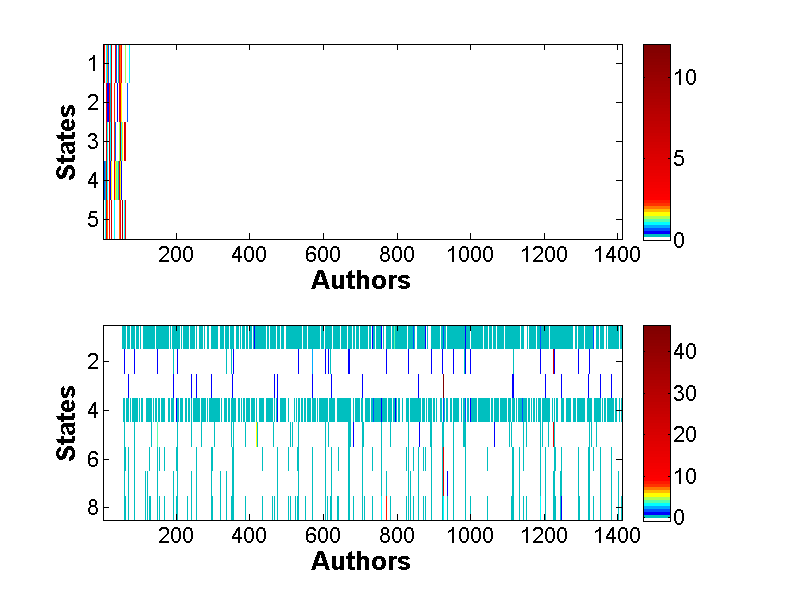}
\caption{Estimated group dictionary for the first group (top plot) and for the second group (bottom plot)}\label{fig:particles_groups}
\end{figure}

\section{Conclusion}
This paper introduced a general modeling framework for learning group dynamics in observed data collected from multiple information sources and over time. The proposed model allows us to analyze relationships between different sources and to make inferences on the temporal patterns of the group behaviors.  

\section{Appendix}
Derivation of Eqn. (\ref{state_filter}).
\begin{align}
&\mathcal{F}({\bs}_{i,j}(t+1))\nonumber\\
=& \sum_{k,m=1}^{Q_j}{\E}_Q\big\{\Lambda_{i}(t+1)\langle {\bs}_{i,j}(t+1), {\be}_{m} \rangle\langle {\bs}_{i,j}(t), {\be}_{k} \rangle| \underline{\mathcal Z}_{t+1}\big\}{\be}_m\nonumber\\
=&\sum_{k,m=1}^{Q_j}\sum_{\boldsymbol \ell \in \mathcal{A}_j(m)}\lambda_i^{\boldsymbol \ell}(t+1){\E}_Q\{\Lambda_{i}(t) \langle {\bA}_j{\bs}_{i,j}(t), {\be}_{m} \rangle \nonumber\\
&\times\langle {\bs}_{i,j}(t), {\be}_{k} \rangle\mid \underline{\mathcal Z}_{t}\}{\be}_m\nonumber\\
=&\sum_{k,m=1}^{Q_j}\sum_{\boldsymbol \ell \in \mathcal{A}_j(m)}\lambda_i^{\boldsymbol \ell}(t+1){\bA}_j(k,m)\langle{\mathcal F}({\bs}_{i,j}(t)), {\be}_{k}\rangle{\be}_{m}.\nonumber
\end{align}

\ifCLASSOPTIONcaptionsoff
  \newpage
\fi

\bibliographystyle{IEEEtran}
\bibliography{Reference}

\end{document}